\documentclass[letterpaper, 10 pt, conference]{ieeeconf}  

\IEEEoverridecommandlockouts                              

\usepackage{amsmath,amsfonts}
\usepackage{algorithmic}
\usepackage{algorithm}
\usepackage{array}
\usepackage[caption=false,font=normalsize,labelfont=sf,textfont=sf]{subfig}
\usepackage{textcomp}
\usepackage{stfloats}
\usepackage{url}
\usepackage{verbatim}
\usepackage{graphicx}
\usepackage{booktabs}
\usepackage{cite}
\usepackage{balance}
\usepackage[T1]{fontenc}
 \usepackage{balance}
\usepackage{xcolor}
\usepackage{tikz}
\usetikzlibrary{calc}
\usepackage{eso-pic}

\title{\LARGE \bf
Performance Evaluation of Transfer Learning Based Medical Image Classification Techniques for Disease Detection
}

\author{Zeeshan Ahmad$^{1}$,~\IEEEmembership{Member,~IEEE,} Shudi Bao$^{1,*}$,~\IEEEmembership{Member,~IEEE,}  and Meng Chen$^{2}$
\thanks{This work was supported in part by the Open Research Fund of National Mobile Communications Research Laboratory, Southeast University (No.2023D15), Ningbo ClinicaI Research Center for Medical Imaging (No.2022LYKFYB01), and the Natural Science Foundation of Zhejiang, China (LGF22H120009). (\it{$^{*}$Corresponding author: Shudi Bao})}
\thanks{$^{1}$Zeeshan Ahmad and Shudi Bao are with the Ningbo Institute of Digital Twin, Eastern Institute of Technology, Ningbo 315200, People's Republic of China
       (email: {\tt\small sdbao@idt.eitech.edu.cn})}%
\thanks{$^{2}$Meng Chen is with the School of Cyber Science and Engineering, Ningbo University of Technology, Ningbo 315211, People's Republic of China
       (email: {\tt\small cm@nbut.edu.cn})}%
}

\begin{document}

\AddToShipoutPictureFG{
    \begin{tikzpicture}[remember picture,overlay]
        \node[anchor=north, yshift=-5pt] 
            at (current page.north) {\textbf{Published as Conference Paper at IEEE EMBC 2025}};
        \draw[line width=0.5pt]
            ($(current page.north west)+(0,-18pt)$) --
            ($(current page.north east)+(0,-18pt)$);
    \end{tikzpicture}
}

\maketitle
\thispagestyle{empty}
\pagestyle{empty}

\begin{abstract}

Medical image classification plays an increasingly vital role in identifying various diseases by classifying medical images, such as X-rays, MRIs and CT scans, into different categories based on their features. In recent years, deep learning techniques have attracted significant attention in medical image classification. However, it is usually infeasible to train an entire large deep learning model from scratch. To address this issue, one of the solutions is the transfer learning (TL) technique, where a pre-trained model is reused for a new task. In this paper, we present a comprehensive analysis of TL techniques for medical image classification using deep convolutional neural networks. We evaluate six pre-trained models (AlexNet, VGG16, ResNet18, ResNet34, ResNet50, and InceptionV3) on a custom chest X-ray dataset for disease detection. The experimental results demonstrate that InceptionV3 consistently outperforms other models across all the standard metrics. The ResNet family shows progressively better performance with increasing depth, whereas VGG16 and AlexNet perform reasonably well but with lower accuracy. In addition, we also conduct uncertainty analysis and runtime comparison to assess the robustness and computational efficiency of these models. Our findings reveal that TL is beneficial in most cases, especially with limited data, but the extent of improvement depends on several factors such as model architecture, dataset size, and domain similarity between source and target tasks. Moreover, we demonstrate that with a well-trained feature extractor, only a lightweight feedforward model is enough to provide efficient prediction. As such, this study contributes to the understanding of TL in medical image classification, and provides insights for selecting appropriate models based on specific requirements.

\end{abstract}

\section{INTRODUCTION}

Medical imaging is a key research domain in modern healthcare systems. It enables a non-invasive visualization of anatomical structures and detailed description of various physiological processes. Various techniques, such as X-ray, computed tomography (CT), magnetic resonance imaging (MRI), and positron emission tomography (PET) are used for the diagnosis of numerous diseases followed by treatment planning and patient monitoring \cite{c1}. Generally, X-rays are used to detect fractures in body parts. Besides, they are also used to identify different diseases, such as pneumonia \cite{c2}, lung cancer \cite{c3}, emphysema, heart problems \cite{c4}, multiple fundus images detection \cite{c5}, and covid detection \cite{c6}. Pneumonia is a condition where the air sacs in the lungs become filled with fluid or air, leading to breathing difficulties. In chest X-rays of individuals with pneumonia, white patches are often visible. Pulmonary nodules, on the other hand, are small spots that can develop in the lungs, and may be either malignant (cancerous) or benign (noncancerous). These nodules also appear as white spots in X-ray images and can be identified using low-dose CT scans.

In recent years, the application of deep learning (DL) models in healthcare has attracted great research interest, with the majority concentrated on medical imaging. It has been verified that neural networks \cite{c7} can generalize any nonlinear function since they follow the universal theorem \cite{c8}. Given a dataset, the neural network aims to find an efficient data-based surrogate model for predicting output features.

At present, there have been several comprehensive and in-depth studies conducted on the application of DL techniques in medical and healthcare domain. For example, Litjens et al. \cite{c9} reviewed over 300 articles on DL for medical image analysis, whereas Chowdhury et al. \cite{c10} examined state-of-the-art research on self-supervised learning in medicine. Other studies have concentrated on transfer learning (TL) with specific case studies, such as microorganism counting \cite{c11}, cervical cytopathology \cite{c12}, neuroimaging biomarkers of Alzheimer’s disease \cite{c13}, and magnetic resonance imaging of the brain. Kim et al. \cite{c14} provided a comprehensive review of TL techniques for medical image classification.

Despite the success of neural networks in recent years, most of the modelling techniques face issues, such as model interpretation \cite{R1}, error convergence \cite{R2}, and class imbalance \cite{R3}. Model interpretation is related to the layerwise parameter analysis. For a deep learning model, such as a convolutional neural network (CNN), the analysis becomes intractable due to its millions or even billions of parameters. In most cases, the training algorithm faces convergence issues which are mostly related to the optimization of a highly nonlinear objective function with possible local minima and saddle points. Class imbalance is another issue which comes from the data distribution. If a dataset has more data of one class “A”, and scarce data for another class “B”, then the model tends to provide more importance for prediction of class “A” compared to class “B”. 

Model performance for training a large neural network from scratch depends on the initial weight distribution. Since initially all the weights are randomly distributed, there always remains an uncertainty in the model performance. TL is a popular method that uses pre-trained models as a starting point for a new task, leveraging knowledge gained from one domain to improve performance in another domain \cite{c14}. This approach is particularly beneficial in medical imaging, where large labeled datasets are often scarce. By utilizing pre-trained models, typically from natural image datasets, e.g., ImageNet, TL can significantly reduce training time, computational costs, and the need for extensive labeled data in medical image classification tasks \cite{c11, c12, c13}. Several studies have demonstrated the effectiveness of TL in medical imaging, such as classification of skin and breast cancer images \cite{c14}, and pneumonia detection in chest X-rays \cite{c15}.

In this paper, we present a comprehensive case study of several state-of-the-art DL models for a custom medical imaging dataset. In particular, we have shown that rather than training a model from scratch, if we consider training a pre-trained model, the convergence of loss function is numerous times faster compared to the earlier approach. In addition, we also provide the uncertainty analysis which provides a bound on the loss value for certain training process, and runtime comparison of different models to provide background into the model’s reproducibility. From this analysis, we will be able to find which type of deep learning architecture is useful for a particular type of dataset. This study focuses on models with convolutional feature extractors. While Vision Transformer (ViT) based models have shown efficiency in medical image analysis, they have fundamentally different architectures than convolutional models. Therefore, ViT based models are are excluded from our comparative analysis.

The rest of the paper is organized as follows. Section II provides a detailed description of the methodology. In section III, we conduct experiments to evaluate the performance of various state-of-the-art TL based deep learning techniques. Finally, we make our conclusions and discuss future work in section IV.


\section{METHODOLOGY}
Neural networks are empirical models that are inspired by and resembles the idea of neural information processing of the nervous systems. The fundamental building block of a neural network is a series of interconnected nodes which constitute a layer. These nodes process the input data from the early layer nodes by applying weights and biases, and computes the final output based on learned features in the output layer.

Given the input matrix $X \in \mathbb{R}^{m, n}$, where $m$ is the number of data points and $n$ is the number of features, the output $z_{i}^{l}$ of the $l$th layer for the $i$th data point is then given by
\begin{equation}
    z_{i}^{(l)} = \sum_{j} w_{ij}^{(l)} a_{j}^{(l-1)} + b_{i}^{(l)}
\label{eq1}
\end{equation}
where $w_{ij}^{(l)}$ and $b_{i}^{(l)}$ respectively denote the weight  and bias associated with the $i$th neuron in the $l$th layer, and 
\begin{equation}
a_{i}^{(l)} = \phi(z_{i}^{(l)})
\label{eq2}
\end{equation}
is the activations from the previous layer $(l-1)$. Here, $\phi$ denotes a nonlinear activation function that introduces non-linearity into the model, and $X$ can be considered as $a_{j}^{(0)}$, that is, the inputs to the first hidden layer. Without a non-linear activation function, the neural network is essentially a linear regression model, which performs a linear transformation from the input space to the output space.

\begin{figure}
    \centering
    \includegraphics[width=1\linewidth]{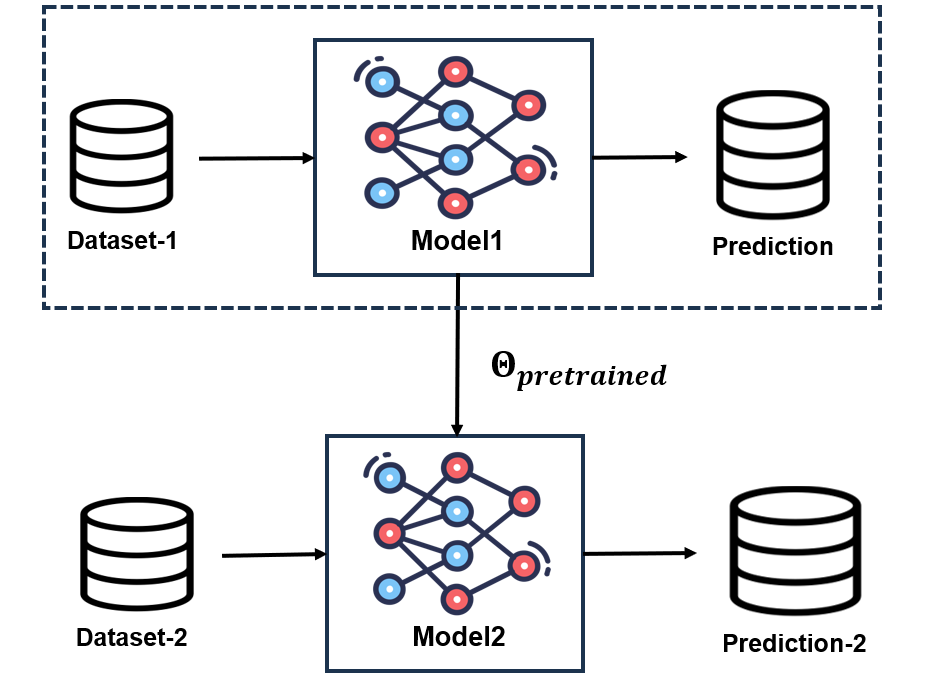}
    \caption{Schematic diagram of transfer learning}
    \label{fig1}
\end{figure}

The concept of TL arises from the ability of lower layers of a deep neural network to capture generic features of the data, such as horizontal and vertical patterns in images. The schematic diagram of a simple TL approach is shown in Fig. \ref{fig1}.

Let consider a DL model $\theta_{source}$ that is trained on a source task, minimizing the source loss $L_{S}$ by updating the weights $\theta$ as shown in Eq. \ref{eq3}. After numerous iterations following the Eq. \ref{eq5}, we learn the weights which is now applied to the target learning task shown in Eq. \ref{eq4}. During training the model for target task, part of the pre-trained network remains fixed (frozen) and only deeper network layers are trained following Eq. \ref{eq6}. Here, $f_{shared}$ denotes the pre-trained network's layers which are frozen, and only task specific layers $f_{specific}$ are trained.
\begin{equation}
\theta_{source} = arg min_{\theta} \frac{1}{N_{S}} \sum_{i=1}^{N_{S}} L_{S}(f_{source}(x_{i}^{T}; \theta), y_{i}^{S})
\label{eq3}
\end{equation}

\begin{equation}
    \theta_{target} = arg min_{\theta} \frac{1}{N_{T}} \sum_{i=1}^{N_{T}} L_{T}(f_{target}(x_{i}^{T}; \theta), y_{i}^{T})
\label{eq4}
\end{equation}

\begin{equation}
\theta_{target}^{(k)} = \theta_{target}^{(k-1)} = \alpha \nabla_{\theta} L_{T}^{(k)}
\label{eq5}
\end{equation}

\begin{equation}
f_{target}(x; \theta) = f_{specific}(f_{shared}(x; \theta_{shared}); \theta_{specific})
\label{eq6}
\end{equation}

We explore different pre-trained models, such as ResNet \cite{c16}, AlexNet \cite{c17}, VGG\cite{c18} and InceptionV3 \cite{c19}, in the comparative study. A majority of these networks considers residual network based models as the building block. Residual networks are efficient in solving vanishing gradient problem and are robust during learning process.
Residual layers can be expressed as
\begin{equation}
H^{(l)}_{i}(x) = [\sum_{j} w_{ij}^{(l)} a_{j}^{(l-1)} + b_{i}^{(l)}] + H^{(l-k)}_{j}(x)
\label{eq7}
\end{equation}
where $w_{ij}^{(l)}$ is the kernel matrix which converts the input space of $j$th dimension to an output space of the $i$th dimension, and $H$ represents the output of a hidden layer. We compute $H_{i}^{(l)}$ for the $i$th node in the $l$th layer using Eq. \ref{eq7}. Since we are concatenating the output of a previous hidden layer to the output of the current hidden layer, the size of both layer's output should be the same. Hence, the length of index $i$ should be same as the length of index $j$. According to this statement, the kernel matrix $w_{ij}$ should be a square matrix. This process is called identity mapping, which provides a context of identity of previous layer to the next layer. Identity mapping improves the training process by mitigating vanishing gradient problems. Besides ResNet, we evaluate InceptionV3 model \cite{c19}, which uses factorized convolution, batch normalization and label smoothing to improve the performance of a highly deep network.

\section{RESULTS AND DISCUSSION}

In this section, we conducted extensive experiments to evaluate the performance of most widely used TL models for medical image classification, including AlexNet, VGG16, different variants of ResNet model, and InceptionV3 model. We consider three variants of the ResNet model: ResNet-18, ResNet-34 and ResNet-50. Using these ResNet variants in the comparison helps to understand the performance variation of models with the depth of a neural network.

\subsection{Exploratory Analysis on the Dataset}

The dataset used to evaluate various TL models consists of chest X-ray images showing pneumonia, tuberculosis, and normal conditions, which is sourced from Kaggle’s publicly available datasets \cite{R4}. We split the dataset into training, validation and testing sets after shuffling the data to provide generalized training. The number of data points in training, validation and testing sets are $13028$, $761$ and $1527$, respectively. In the training set, we have maintained identical distribution among different classes. For example, normal, pneumonia, tuberculosis and unknown classes in the training set have $4667$, $3633$, $3573$ and $1155$ data points, respectively. The unknown class contains images where there are some anomalies, and the disease condition is not well represented. Rather than discarding these images from the dataset, we have included them as a separate class. This enhances the model’s ability to distinguish anomalous conditions from both normal and known disease states. The original images in the dataset had a shape of (1272, 1040), which were resized to $(224, 224)$ during modelling for computational efficiency. Additionally, we converted the dataset to grayscale before modelling, which has only one channel. Besides these primary transformations, Random Horizontal Flip, Random Rotation, and Normalization were also applied to provide diversity to the data distribution and increase the model robustness to external perturbations.
\begin{table}[]
  \caption{Performance comparison of various models on training set}
\centering
  \label{tab1}
\begin{tabular}{@{}lllllll@{}}
\toprule
Model       & Precision                 & Recall                                            & Accuracy                                                                 & F1 Score   \\ \midrule
AlexNet    & {0.931} & {0.938} & {0.926} & {0.93} \\
VGG16    & {0.957} & {0.961} & {0.957} & {0.951} \\
ResNet18    & {0.976} & {0.961} & {0.972} & {0.968} \\
ResNet34    & {0.983} & {0.98}  & {0.974} & {0.979} \\
ResNet50    & {0.992} & {0.984} & {0.981} & {0.99}  \\
InceptionV3 & {0.994} & {0.989} & {0.99}  & {0.991} \\
\bottomrule
\end{tabular}
\end{table}

\begin{table}[]
  \caption{Performance comparison of various models on testing set}
\centering
  \label{tab2}
\begin{tabular}{@{}lllllll@{}}
\toprule
Model       & Precision                 & Recall                                            & Accuracy                                                                & F1 Score \\ \midrule
AlexNet    & {0.922} & {0.918} & {0.926} & {0.927} \\
VGG16    & {0.935} & {0.93} & {0.941} & {0.939} \\
ResNet18    & {0.961} & {0.955} & {0.964} & {0.952} \\
ResNet34    & {0.977} & {0.962} & {0.968} & {0.97}  \\
ResNet50    & {0.98}  & {0.976} & {0.97}  & {0.98}  \\
InceptionV3 & {0.987} & {0.981} & {0.989} & {0.985} \\ \bottomrule
\end{tabular}
\end{table}

\subsection{Experiment Configuration}
For training the models, we first froze the initial layers of the feature extractor and trained the final set of layers. Since we are using TL, it is assumed that initial layers are capable of extracting generic features. For training each model, we used Adam Optimizer \cite{c21} with a learning rate of $0.001$, and a batch size of $32$. All the models in the case study are convolutional models. The ResNet models have a common hyperparameter configuration. The kernel size for each residual model in the core feature extraction block is set to $(3, 3)$, whereas the global average pooling layers use a kernel size of $(7, 7)$. These architectures were pre-trained on ImageNet dataset and stored as ``.pt" files in pytorch framework. Similarly, for Inception model building blocks, the kernel size and strides are set to $(3, 3)$ and $(2, 2)$, respectively. Inception model concatenates the output from multiscale transformation of the input and creates an output with different types of features. Additionally, we also analyzed the parameter pool size of each model. ResNet-18, ResNet-34, ResNet-50, and InceptionV3 model have 11.18M, 21.3M, 23.5M and 25.12M parameters, respectively. With such a huge number of parameters, they are highly computationally intensive and time consuming to train. TL freezes the pre-trained features extractor, significantly saving time and resource of intensive training. We used a global average pooling layer to aggregate features from the pre-trained model, which are then passed through a feedforward network (FFN) for classification. The number of hidden nodes follow the sequence $512-128-64-64-2$. Since a standard FFN is used for all models, we fixed its structure to ensure consistency and avoid ambiguity in comparisons.

Python was used for code implementation. The models were trained on Google Colab using an NVIDIA T4 GPU system. We implemented the code using PyTorch libraries with additional analysis performed using NumPy, Pandas, Matplotlib, and other supporting libraries. 

\begin{figure*}
    \centering
    \includegraphics[width=0.8\linewidth, height=7cm]{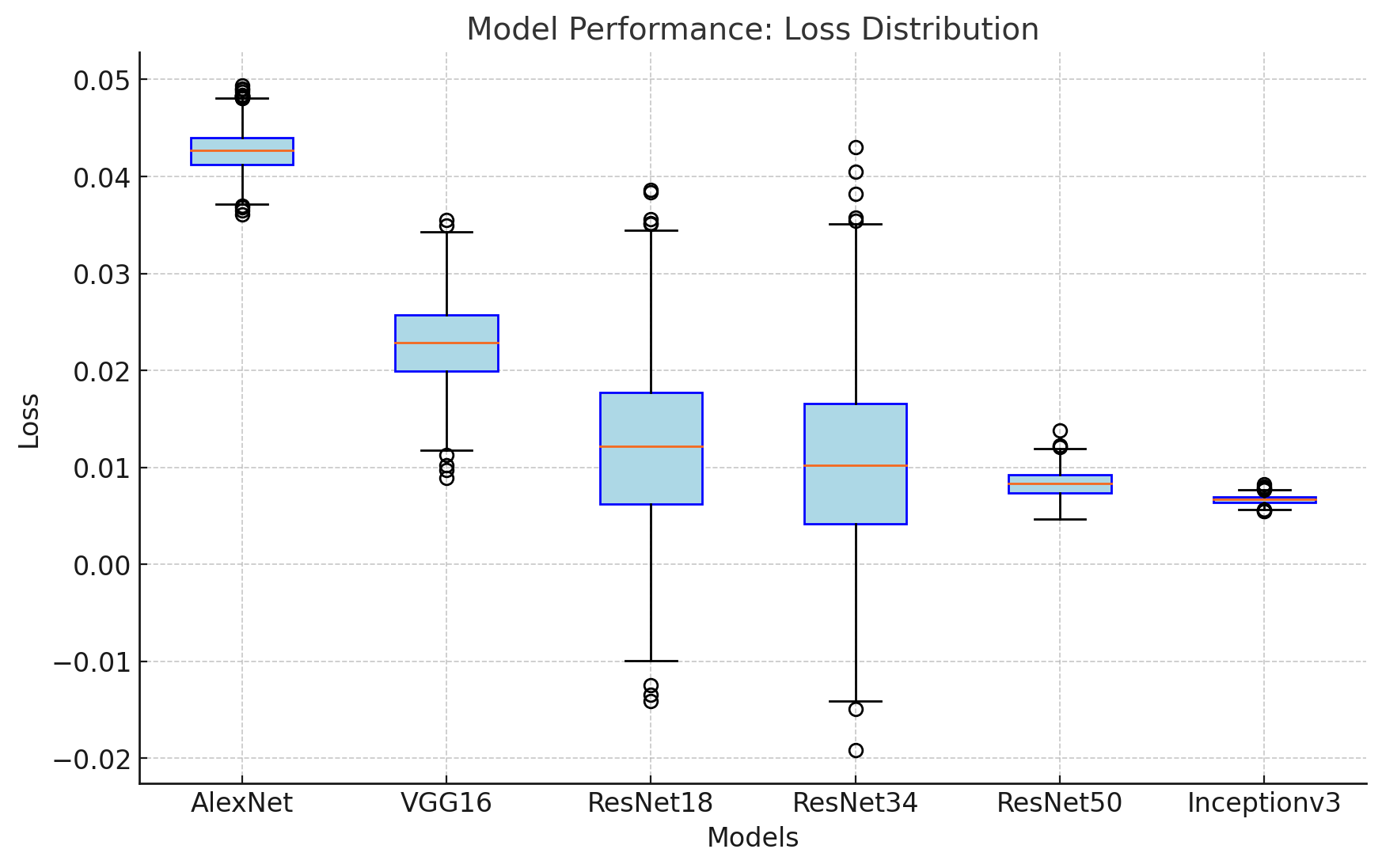}
    \caption{Performance variations of various models}
    \label{fig2}
\end{figure*}

\subsection{Evaluation Metrics}

We evaluate the prediction performance of different models for classification of medical images using well established metrics: Precision, Recall, Accuracy, and F1 Score. These evaluation metrics are derived from the confusion matrix, consisting of True Positives (TP), True Negatives (TN), False Positives (FP), and False Negatives (FN). Each metric reflects a specific aspect of the model's predictive capability.

Precision measures the proportion of correctly identified positive cases among all the predicted positive cases. A higher precision value indicates fewer FP and is crucial for applications where minimizing incorrect positive predictions is essential, such as in disease diagnosis. Recall, on the other hand, quantifies the proportion of actual positive cases correctly identified by the model. It is vital in scenarios where missing positive instances (FN) can have significant consequences, such as failing to detect a critical medical condition. Accuracy provides an overall measure of the model's performance by calculating the proportion of correctly classified cases, both positive and negative, out of the total number of cases. While accuracy offers a comprehensive view, it may be less reliable in the presence of imbalanced datasets. The F1 Score serves as the harmonic mean of Precision and Recall, balancing the trade-off between these two metrics. This metric is particularly beneficial for datasets with imbalanced classes, as it ensures that neither precision nor recall is disproportionately favored.

\subsection{Model Comparison Analysis}
The results of training and testing performance of various TL models are presented in Tables \ref{tab1} and \ref{tab2}, respectively. It can be observed from Tables \ref{tab1} and \ref{tab2} that InceptionV3 outperforms other models across all the metrics, achieving highest Precision, Recall, Accuracy, and F1 Score during both the training and testing phases. Specifically, InceptionV3 achieves a testing precision of 0.987, recall of 0.981, accuracy of 0.989, and F1 Score of 0.985. This suggests that InceptionV3 excels in maintaining a balance between minimizing false positives and false negatives, making it highly suitable for medical image classification task. ResNet50 also demonstrated strong performance, with a testing precision of 0.980, recall of 0.976, accuracy of 0.970, and F1 Score of 0.980, showing its effectiveness in accurately classifying medical images. Among remaining models, ResNet34 shows better performance than ResNet18, with higher values for precision, recall, accuracy, and F1 Score. ResNet34 achieves a testing precision of 0.977 and an F1 Score of 0.970, whereas ResNet18 attains slightly lower values (precision: 0.961, F1 Score: 0.952) during testing. These results indicate that deeper architectures, such as ResNet50 and InceptionV3, are more capable of capturing complex patterns in medical images compared to shallower architectures. VGG16 and AlexNet, though performing reasonably well, show lower accuracy. This difference highlights the rapid progress in DL model design and the benefits of more sophisticated architectures for medical image analysis.

\subsection{Uncertainty Analysis}

To assess the robustness and reliability of these comparative models, we conducted an uncertainty analysis during the validation phase. We considered multiple random splits of the model and repeated the same analysis numerous times for each realization of train-validation-test set to observe how the model is affected by the inherent variability of the neural network architecture. The results are illustrated in Fig. \ref{fig2}. InceptionV3 demonstrates the lowest mean loss (0.0067) and standard deviation of loss (0.0004), indicating not only superior performance but also high consistency and reliability. The ResNet models shows progressively lower mean loss and standard deviation as the depth increased, consistent with their performance metrics. Interestingly, while AlexNet and VGG16 have higher mean losses, they show relatively low standard deviations, suggesting consistent (albeit lower) performance across validation runs.

\subsection{Runtime Comparison}

\begin{figure}
    \centering
    \includegraphics[width=1\linewidth]{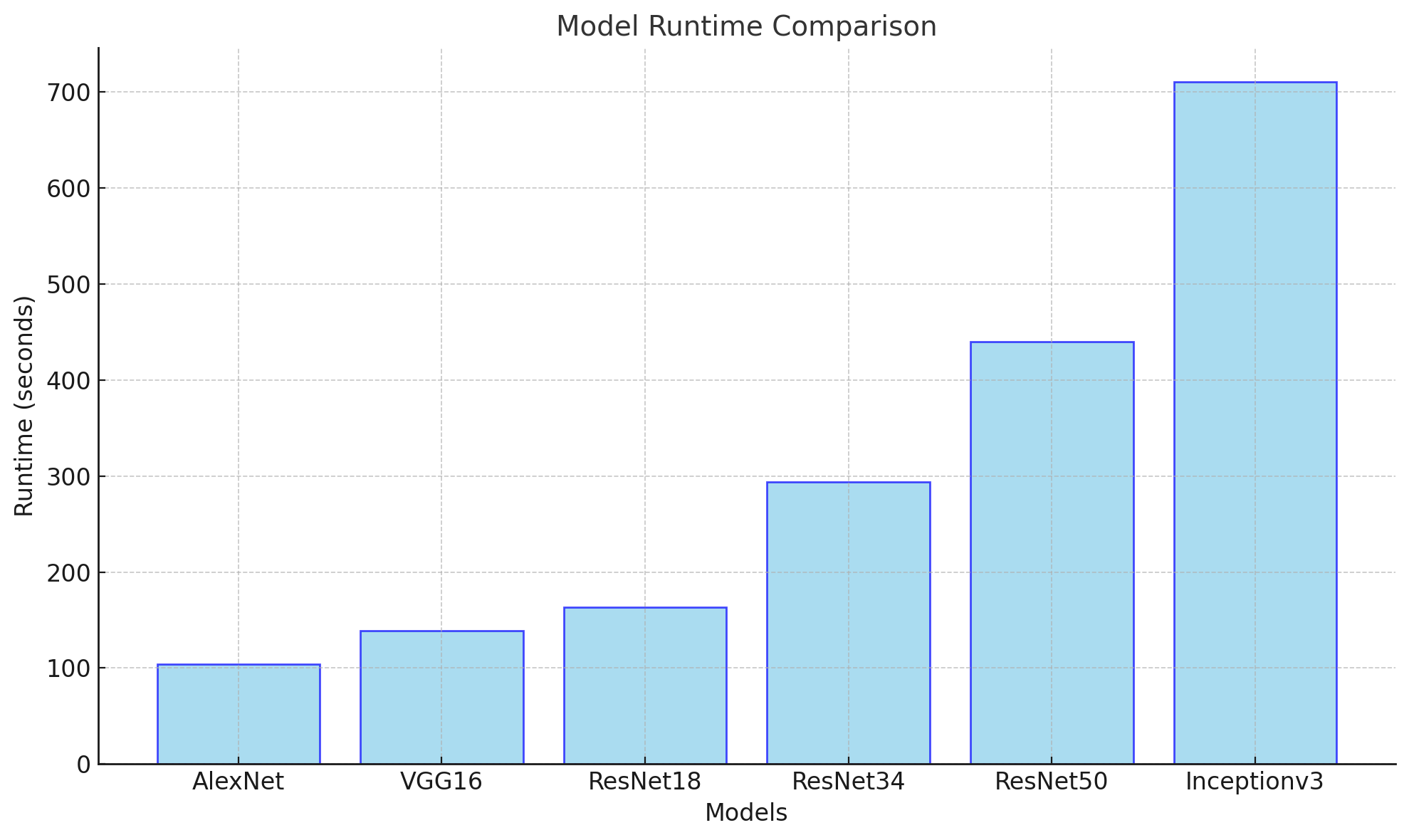}
    \caption{Runtime comparison}
    \label{fig3}
\end{figure}

We also evaluate the computational efficiency of each model by measuring their validation runtime. The results are visualized in Fig. \ref{fig3}. The runtime analysis reveals a trade-off between model performance and complexity. InceptionV3, while providing the best performance, also requires the longest runtime (711 seconds). In contrast, simpler models like AlexNet and VGG16 are significantly faster but at the cost of lower accuracy. The ResNet family shows a clear correlation between depth and runtime, with ResNet50 taking nearly three times as long as ResNet18. This relationship highlights the need to balance model performance with available computational resources in practical applications.

\section{CONCLUSIONS AND FUTURE WORK}

In this paper, we evaluate the performance of the most widely used TL models for medical image classification. TL offers numerous advantages in medical image analysis, including improved convergence, faster training, computational efficiency, generalization ability, and address class imbalance and data scarcity. Experimental results reveal that InceptionV3 offers the best overall performance in terms of accuracy, precision, recall, and F1 score. It also demonstrates the highest consistency and reliability as evidenced by its low uncertainty metrics. However, this superior performance comes at the cost of increased computational requirements. The ResNet family of models, particularly ResNet50, offers a good balance between performance and computational efficiency, making them suitable for a wide range of applications. AlexNet and VGG16, though less accurate, may still be viable options in scenarios where computational resources are limited and slightly lower accuracy is acceptable.

These findings provide valuable insights for researchers in the field of medical image analysis, aiding model selection based on specific needs for accuracy, reliability, and computational efficiency. Future work should explore optimization techniques for complex models to achieve faster inference with minimal accuracy loss, enhancing their viability in resource-constrained environments.

\addtolength{\textheight}{-12cm}   


\balance

\end{document}